\documentclass{article}

\usepackage[preprint]{neurips_2026}

\usepackage[utf8]{inputenc} 
\usepackage[T1]{fontenc}    
\usepackage{hyperref}       
\usepackage{url}            
\usepackage{booktabs}       
\usepackage{amsfonts}       
\usepackage{nicefrac}       
\usepackage{microtype}      
\usepackage{xcolor}         
\usepackage{multirow}
\usepackage{graphicx}
\usepackage{wrapfig}
\usepackage{makecell}
\usepackage{pifont}
\usepackage{caption}
\usepackage{threeparttable}
\usepackage{amsmath}
\usepackage{subcaption}

\begin{document}

\title{C3VD-DEFCOL: A Deformable Colonoscopy Dataset with Time-Resolved 3D Ground Truth and Realistic Appearance}

%

\author{%
  \textbf{Ethan Luk}$^{1,*}$, 
  \textbf{Mayank V. Golhar}$^{1,*}$, 
  \textbf{Anthony Song}$^1$, 
  \textbf{Raúl Iranzo}$^2$, \\
  \textbf{Víctor M. Batlle}$^2$, 
  \textbf{Lalithkumar Seenivasan}$^3$, 
  \textbf{José M.M. Montiel}$^2$, 
  \textbf{Nicholas J. Durr}$^{1, \dagger}$ \\[2ex]
  $^1$Department of Biomedical Engineering, Johns Hopkins University, Baltimore, MD, USA \\
  $^2$Instituto de Investigación en Ingeniería de Aragón (I3A), Universidad de Zaragoza, Zaragoza, Spain \\
  $^3$Department of Computer Science, Johns Hopkins University, Baltimore, MD, USA \\
  \texttt{ndurr@jhu.edu}$^{\dagger}$
}
\maketitle
\def\thefootnote{*}\footnotetext{Equal contribution.}
\def\thefootnote{$\dagger$}\footnotetext{Corresponding author.}
\def\thefootnote{\arabic{footnote}} 

\begin{abstract}
  3D reconstruction could improve colonoscopy by estimating mucosal coverage and alerting clinicians to missed regions during screening. However, algorithm development is limited as no current datasets provide both a realistic in vivo appearance and dense, time-resolved 3D ground truth, especially under non-rigid deformation. We present C3VD-DEFCOL, a framework and dataset for evaluating deformable colonoscopy reconstruction with paired geometry and realistic texture. Starting from C3VD/C3VDv2 colon meshes and camera trajectories, we generate controlled deformations of the colon surface, including peristaltic waves and centerline motion, and render per-frame depth, surface normals, optical flow, occlusion masks, camera poses, and time-stamped 3D meshes. We then use the rendered geometry, primarily depth, to condition an LTX-2.3-based sim-to-real translation model that produces RGB clips with in vivo-like mucosal color, texture, vasculature, and specular appearance while preserving the underlying 3D scene structure. The resulting dataset contains 110 videos from 11 unique colon mesh geometries, with varying camera trajectories, appearances, and parameterized deformation regimes, including three peristaltic severity levels that serve as controlled evaluation axes. We evaluate the generated videos using appearance realism, geometric consistency, and temporal consistency metrics, and use the paired ground truth to benchmark the downstream task of pose estimation in deformable 3D reconstruction. Our experiments show how pose estimation error increases with increasing deformation severity, providing a controlled stress test that is not possible with existing in vivo datasets. Overall, C3VD-DEFCOL is designed as a reproducible, quantitative evaluation platform for testing deformable 3D reconstruction algorithms, with the goal of reducing the domain gap between synthetic datasets and in vivo colonoscopy.
\end{abstract}

\section{Introduction}
Colorectal cancer is the second leading cause of cancer-related mortality in the United States \cite{siegel2025cancer}. Colonoscopy is the gold-standard procedure for screening and removal of precancerous lesions. However, the effectiveness of colonoscopy depends on complete visualization of the mucosal surface. Prior work reports that approximately 26\% of adenomas may be missed during colonoscopy \cite{zhao2019magnitude}, possibly because a 20\% fraction of the colon surface can remain unobserved during routine procedures \cite{mcgill2021artificial}. 3D reconstruction of the colon could reveal missed regions as gaps in the reconstructed colon surface \cite{ma2021rnnslam} and can be used to quantify mucosal coverage and guide re-inspection, potentially leading to a higher adenoma detection rate.

While a broad body of work has studied 2D colonoscopy image analysis, including lesion classification and vascular analysis \cite{golhar2020improving,golhar2024gan,golhar2017robust,golhar2018blood}, 3D reconstruction in colonoscopy remains difficult despite its potential role in quantifying mucosal coverage. Colonoscopy videos are acquired within a deformable, mucous-covered tubular organ under challenging conditions, including weak and repetitive texture, specular highlights, debris, and motion blur \cite{ali2021deep}. However, existing datasets rarely combine these visual challenges while providing dense quantitative geometric ground truth for evaluation. Real in vivo datasets \cite{azagra2023endomapper} capture realistic appearance and artifacts, but generally lack dense pixel-wise geometric ground truth. Synthetic \cite{rau2024simcol3d, singh2025simintestine, smolak2026colonsplat} or phantom datasets \cite{bobrow2023colonoscopy, golhar2026c3vdv2} can provide dense depth, pose, normals, optical flow, and 3D models, but they suffer from domain mismatch in appearance and artifacts. When deformation is present in synthetic datasets, dense time-resolved geometric ground truth is typically unavailable or limited. Consequently, algorithms are often evaluated either qualitatively on real videos or quantitatively on data that does not fully reproduce the visual and non-rigid conditions of colonoscopy. This limits the field’s ability to measure robustness under clinically relevant failure modes, especially deformation.

We introduce C3VD-DEFCOL, a deformable colonoscopy video benchmark designed to evaluate 3D reconstruction under controlled non-rigid deformation while retaining realistic endoscopic appearance. C3VD-DEFCOL applies parameterized deformation models to colon meshes derived from C3VD/C3VDv2 \cite{bobrow2023colonoscopy, golhar2026c3vdv2}, renders dense geometric supervision using a fisheye colonoscope camera model, and translates the resulting videos into in-vivo-style appearance using a geometry-conditioned video generation model. The released dataset includes sim2real RGB videos, depth, surface normals, optical flow, camera poses, and per-timestep deforming mesh across variation in colon geometry, camera trajectory, deformation type, and deformation severity.
\begin{figure}
    \centering
    \includegraphics[width=0.9\linewidth]{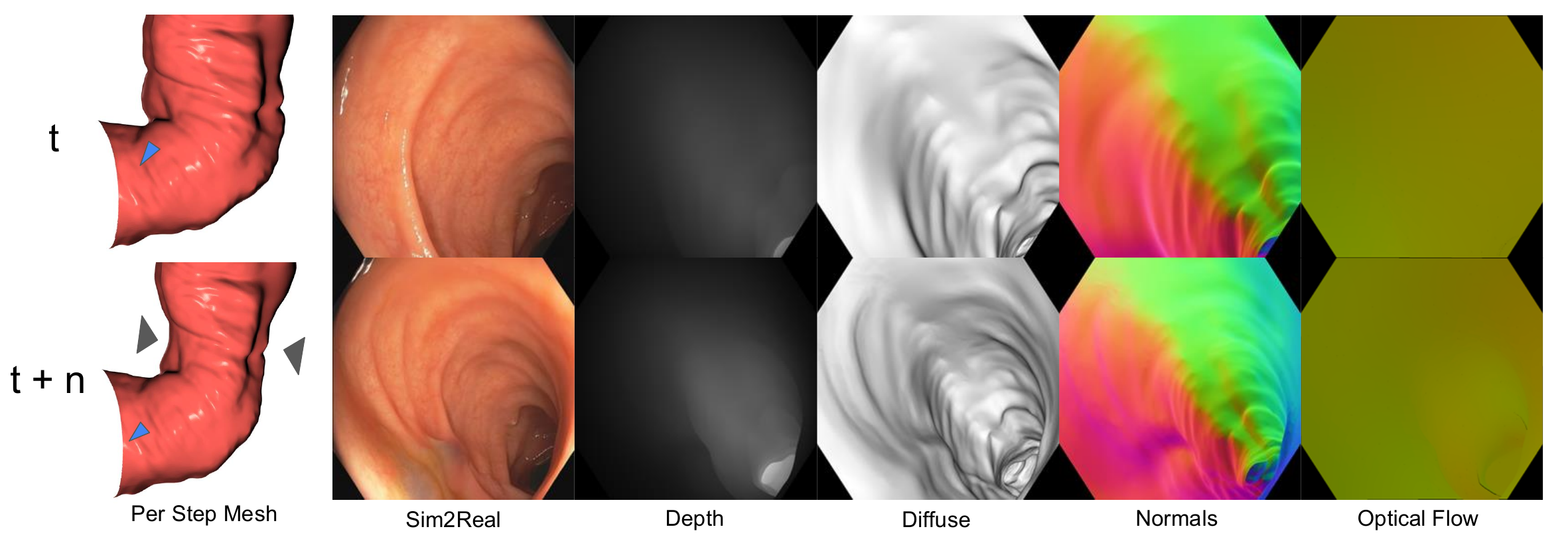}
    \caption{\textbf{C3VD-DEFCOL} provides realistic RGB video and dense ground truth for deformable colonoscopy scenes. For each time instance, we show sim-to-real RGB frames, depth, diffuse, optical flow, surface normals, time-stamped deforming meshes, and the associated camera trajectory, to illustrate the dataset's paired appearance and 3D geometry annotations.}
    \label{fig:overview}
\end{figure}
C3VD-DEFCOL is intended as an evaluation framework that enables controlled, quantitative stress testing of colonoscopy reconstruction algorithms under non-rigid deformation while maintaining realistic video appearance, including colonoscopic artifacts such as vasculature and polyps.  By providing deformation magnitude as an explicit evaluation axis, C3VD-DEFCOL allows researchers to measure how reconstruction error changes as the scene departs from the static-colon assumption. In our experiments, we validate the dataset along three axes: visual realism, geometric consistency between the generated RGB and the known 3D scene, and temporal consistency. We further demonstrate its utility as a benchmark by evaluating camera pose estimation for 3D reconstruction and showing that error increases with increasing deformation severity.

Our contributions are: \textbf{1) A deformable colonoscopy benchmark with dense time-resolved ground truth.} C3VD-DEFCOL provides in-vivo-style RGB videos paired with depth, surface normals, optical flow, camera poses, and per-timestep deforming meshes across controlled deformation types and magnitudes as shown in Figure~\ref{fig:overview}.
\textbf{2) A geometry-conditioned sim-to-real video generation pipeline for colonoscopy.} We adapt an LTX-based video diffusion model with depth conditioning to generate a realistic colonoscopy appearance while preserving consistency with known deforming geometry.
\textbf{3) An open and extensible dataset-generation framework.} The framework allows researchers to generate additional videos with user-specified colon geometry, camera trajectory, deformation type, and magnitude, enabling reproducible evaluation under targeted failure conditions.

\begin{figure}
    \centering
    \includegraphics[width=0.9\linewidth]{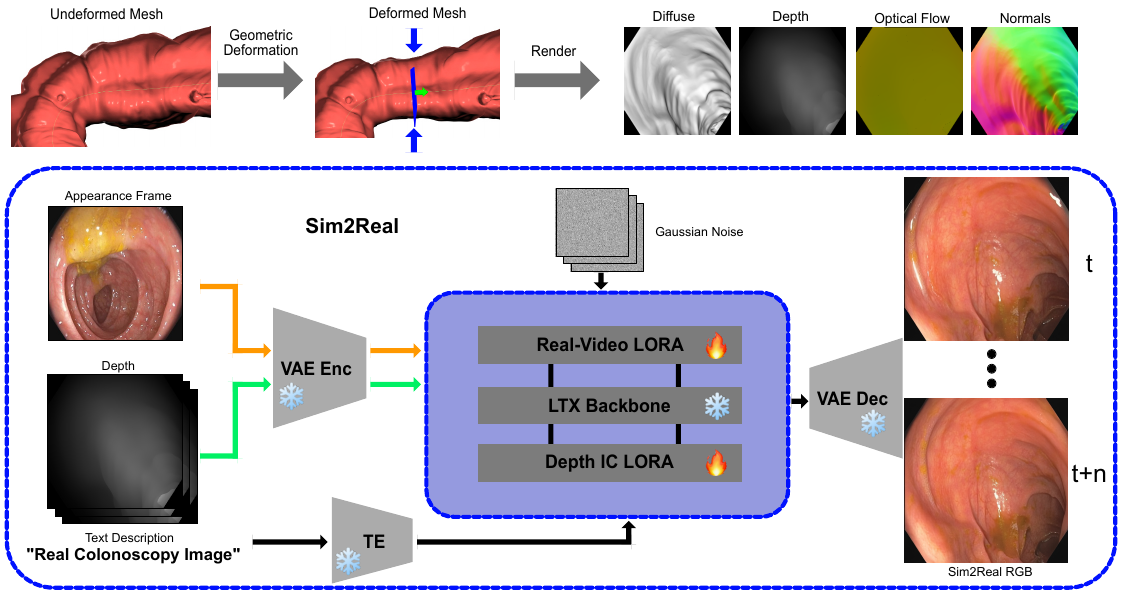}
    \caption{\textbf{C3VD-DEFCOL framework:} Our framework first applies controlled geometric deformation to an undeformed colon mesh and renders the resulting deformable sequence to obtain dense ground-truth. The rendered depth sequence is then used to condition an LTX-based sim-to-real video model, together with a real colonoscopy appearance frame and text prompt, to generate realistic RGB videos that remain aligned with the time-resolved geometry.}
    \label{fig:data_pipeline}
\end{figure}

\section{Related work}

\begin{figure}
    \centering
    \includegraphics[width=0.95\linewidth]{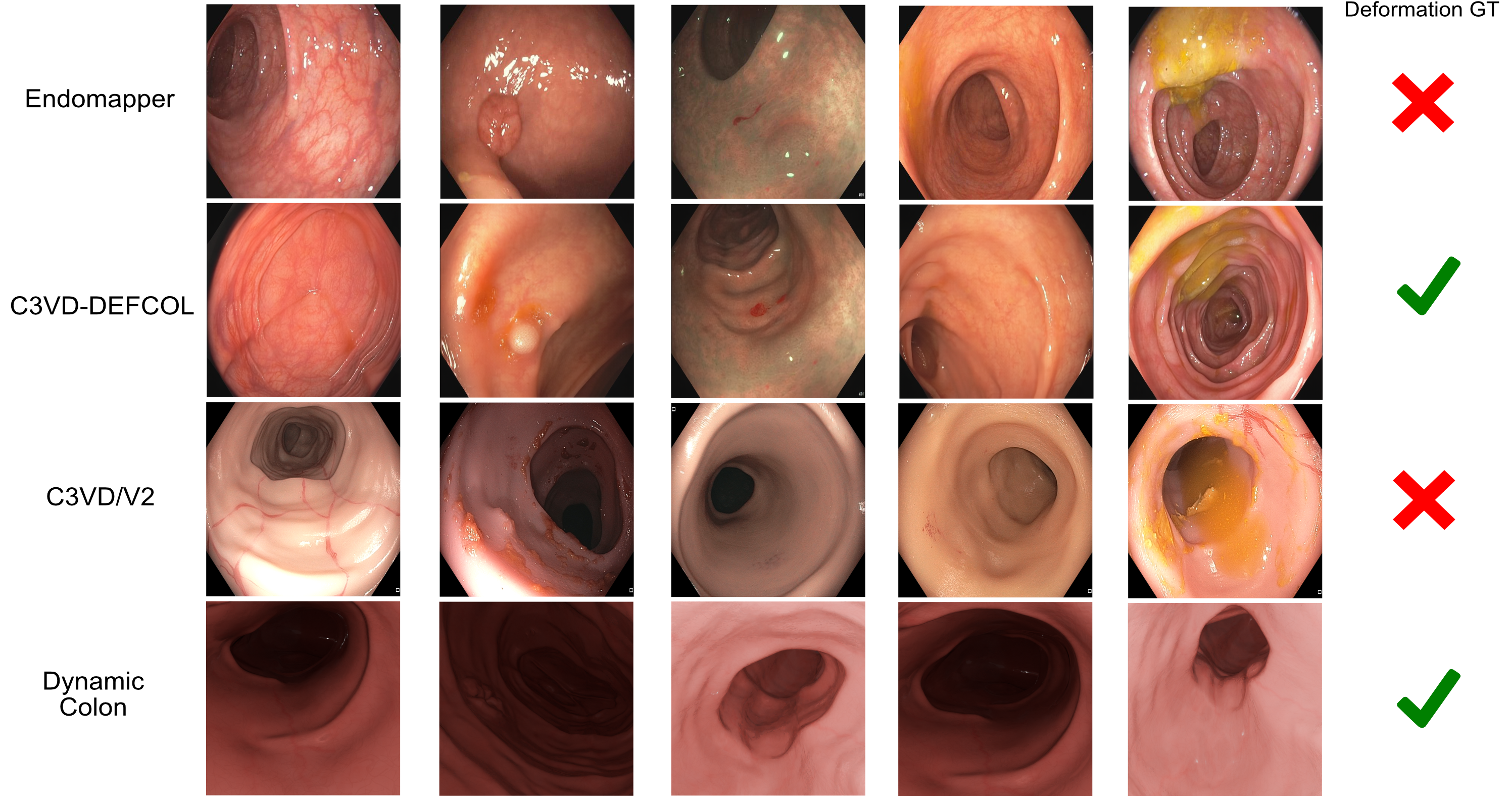}
    \caption{Qualitative appearance comparison. C3VD-DEFCOL sim-to-real RGB images exhibit visual characteristics closer to real in vivo colonoscopy images from EndoMapper while retaining the geometric controllability of synthetic data. Compared with prior synthetic datasets, C3VD-DEFCOL better captures colonoscopy-specific appearance cues such as vascular patterns, polyp contrast, illumination modes, and debris artifacts.}
    \label{fig:realism_compare_datasets}
\end{figure}

\paragraph{3D reconstruction datasets in colonoscopy.} Currently, it is not possible to obtain 3D ground truth for in vivo colonoscopy. So, datasets for colonoscopy 3D reconstruction have typically traded off dense ground truth against realistic appearance as shown in Table \ref{tab:dataset_comparison}. Endomapper \cite{azagra2023endomapper} provides a large collection of calibrated, complete real endoscopy procedures and is valuable for studying V-SLAM \cite{rodriguez2024nr} under realistic clinical conditions, but it lacks dense geometric ground truth. EndoSLAM \cite{ozyoruk2020} acquires ex vivo porcine, phantom, and synthetic data; the synthetic portion has dense per-pixel GT, but the real and ex vivo portions do not. SimCol3D \cite{rau2024simcol3d} established a benchmark for monocular colonoscopy depth and pose prediction using synthetic and real procedure sequences, but the benchmark does not close the appearance gap between synthetic renderings and in vivo colonoscopy. C3VD \cite{bobrow2023colonoscopy} provides clinical-colonoscope phantom videos with registered depth, normals, optical flow, pose, coverage maps, and 3D models, while C3VDv2 \cite{golhar2026c3vdv2} expands scale, artifacts, trajectories, and realism. However, C3VD is static, and C3VDv2 deformation videos lack dense per-step pixel-wise ground truth, leaving quantitative evaluation under non-rigid deformation unresolved.
Several virtual and dynamic datasets address related problems but differ from the conventional colonoscopy setting targeted here. VR-Caps \cite{incetan2021vr} and SimIntestine \cite{singh2025simintestine} simulate wireless capsule endoscopy and include deformation or peristaltic motion with pose/depth supervision, but capsule endoscopy differs from conventional colonoscopy in camera control, optics, illumination, motion statistics, and clinical workflow. While DynamicColon \cite{smolak2026colonsplat}, offers synthetic dynamic colon sequences featuring ground-truth point clouds at each timestep to aid reconstruction under peristalsis, its RGB images are synthetic renderings. It lacks the integrated combination of dense rendered labels and realistic sim-to-real in-vivo RGB appearance; furthermore, its scope is currently restricted to only three video sequences. Illustrative examples are shown in Figure \ref{fig:realism_compare_datasets}. C3VD-DEFCOL fills this gap by combining controllable deforming colon geometry, pixel-wise per-frame geometric ground truth, fisheye colonoscope rendering, and realistic video appearance in a single benchmark for quantitative deformable colonoscopy reconstruction.

\begin{table*}[t]
\centering
\caption{
Comparison of colonoscopy and GI endoscopy datasets for deformable 3D reconstruction evaluation.
C3VD-DEFCOL targets the missing setting of realistic colonoscopy appearance with time-resolved deformable ground truth under controlled deformation.
}
\label{tab:dataset_comparison}
\footnotesize
\setlength{\tabcolsep}{4.5pt}
\renewcommand{\arraystretch}{1.05}

\newcommand{\cmark}{\textcolor{green}{\ding{51}}}
\newcommand{\xmark}{\textcolor{red}{\ding{55}}}
\newcommand{\pmark}{\textcolor{orange!85!black}{$\sim$}}

\resizebox{\textwidth}{!}{
\begin{tabular}{l l l c c c r}
\toprule
\textbf{Dataset} &
\textbf{Camera type} &
\textbf{Tissue type} &
\textbf{\makecell{Realistic\\RGB}} &
\textbf{\makecell{Time-resolved\\deformable GT}} &
\textbf{\makecell{Controlled\\deformation}} &
\textbf{\# Frames} \\
\midrule

EndoSLAM~\cite{ozyoruk2020} &
USB endoscope &
Ex-vivo porcine &
\cmark & \xmark & \xmark &
39,406 \\

EndoSLAM~\cite{ozyoruk2020} &
Pill camera &
Ex-vivo porcine &
\cmark & \xmark & \xmark &
3,294 \\

EndoSLAM~\cite{ozyoruk2020} &
Colonoscope &
Phantom &
\pmark & \xmark & \xmark &
12,250 \\

EndoSLAM~\cite{ozyoruk2020} &
Rendered &
Virtual GI Tract &
\xmark & \xmark & \xmark &
21,887 \\

EndoMapper~\cite{azagra2023endomapper} &
Colonoscope &
In vivo human &
\cmark & \xmark & \xmark &
24 Hrs video\\

SimCol3D~\cite{rau2024simcol3d} &
Rendered Colonoscope&
Virtual Colon&
\xmark & \xmark & \xmark &
37,800 \\

C3VD~\cite{bobrow2023colonoscopy} &
Colonoscope &
Phantom &
\pmark & \xmark & \xmark &
10,015 \\

C3VDv2 (pixel-wise GT)~\cite{golhar2026c3vdv2} &
Colonoscope &
Phantom &
\pmark & \xmark & \xmark &
67,886 \\

C3VDv2 (deformation)~\cite{golhar2026c3vdv2} &
Colonoscope &
Phantom &
\pmark & \xmark & \xmark &
6,185 \\

VR-Caps~\cite{incetan2021vr} &
Rendered capsule &
Virtual GI tract &
\xmark & \cmark & \cmark &
8,714+ \\

SimIntestine~\cite{singh2025simintestine} &
Rendered capsule &
Virtual GI tract &
\xmark & \cmark & \cmark &
8,715 \\

DynamicColon~\cite{smolak2026colonsplat} &
Rendered colonoscope &
Virtual colon &
\xmark & \cmark & \cmark &
600 \\

\textbf{C3VD-DEFCOL (ours)} &
\textbf{Rendered colonoscope} &
\textbf{Virtual colon} &
\cmark & \cmark & \cmark &
\textbf{9,240} \\

\bottomrule
\end{tabular}
}

\vspace{2pt}
\begin{flushleft}
\scriptsize
\cmark: provided; \xmark: not provided; \pmark: partial, subset-specific, or limited comparability.
\textbf{Time-resolved deformable GT} denotes per-frame or per-timestep geometry for a non-rigid scene, not merely a static 3D model or camera pose.
\textbf{Controlled deformation} denotes user-specified or parameterized deformation suitable as an evaluation axis.
\end{flushleft}
\end{table*}

\paragraph{SLAM in colonoscopy.} Existing research in 3D colonic reconstruction has transitioned from classical Structure-from-Motion (SfM) \cite{schonberger2016structure} and V-SLAM systems \cite{rodriguez2024nr} to more sophisticated deep-learning-boosted approaches \cite{ma2021rnnslam,liu2022sage,recasens2024drunkard}. Furthermore, modern neural representations and 3D Gaussian Splatting (3DGS) methods \cite{bonilla2024gaussian,wang2024endogslam,shan2024enerf} show promising results for capturing static scenes and, more recently, non-rigid peristaltic motion but from known camera poses \cite{smolak2026colonsplat}. Non-rigid SLAM systems have been proposed to decouple ego-motion from surface deformation \cite{gomez2021sd,rodriguez2024nr}; however, robust reconstruction in deforming colonoscopy scenes remains a persistent challenge, highlighting the long-standing need for a dataset for rigorous evaluation and benchmarking that simultaneously provides dense ground truth and realistic imagery in deforming scenes.

\section{C3VD-DEFCOL Dataset}

C3VD-DEFCOL is a curated set of 110 synthesized colonoscopy videos with realistic in-vivo-style appearance and pixel-wise, time-resolved geometric ground truth. The dataset spans 11 unique colon segment geometries covering the cecum, ascending, transverse, descending, and sigmoid colon, and includes five motion regimes: rigid, low-, medium-, and high-magnitude peristalsis, and centerline-shift deformation. Each video contains 84 frames at 16 fps and a resolution of $512 \times 512$, yielding 9,240 RGB frames in total. For each frame, we provide sim2real RGB, camera pose, depth, surface normals, optical flow, and per-step mesh information. The deforming surface is represented by a canonical mesh together with per-frame vertex positions and vertex normals. The project page is \url{https://github.com/DurrLab/C3VDv3}.

We generate the dataset in two stages as shown in Figure \ref{fig:data_pipeline}. First, C3VD/C3VDv2-derived colon meshes are deformed using parameterized motion fields and rendered with a calibrated fisheye colonoscope camera model to produce dense geometric supervision. Second, the rendered geometry-conditioned videos are translated into a realistic colonoscopy appearance using a video-to-video diffusion model conditioned on the depth sequence and a reference appearance frame. Up to two appearance variants are generated for each source clip and retained after filtering for visual quality, and geometric consistency.

The motion regimes enable graduated reconstruction evaluation. The rigid setting provides a static-scene baseline for rigid SfM/SLAM methods. The low, medium, and high peristalsis settings progressively increase non-rigid tissue motion while keeping the scene and camera trajectory fixed, allowing performance degradation to be measured as a function of deformation severity. The centerline-shift setting tests robustness to a distinct global deformation pattern. It could test the algorithm's ability to disentangle camera motion from colon deformation. Together, these settings make C3VD-DEFCOL suitable for controlled quantitative evaluation of both rigid and non-rigid colonoscopy reconstruction methods.

\subsection{Deformation generation}

\paragraph{Peristalsis.}
We model peristaltic deformation as a parameterized traveling Gaussian contraction field. During animation, each mesh vertex is projected onto the anatomical centerline, which was computed using the VMTK toolkit in 3D Slicer. The radial offset of each vertex relative to the centerline is defined as \(\mathbf{r}_i = \mathbf{x}_i - \mathbf{c}(s_i)\), 
where \(\mathbf{c}(s_i)\) is the corresponding point on the centerline at arclength position \(s_i\). 

Each contraction wave \(w\) is modeled as a traveling Gaussian:
\begin{equation}
g_{i,w}(t) = A_w \cdot f_w(t) \cdot \exp\left(-\frac{(s_i - v \cdot t)^2}{2\sigma_w^2}\right),
\end{equation}
where \(A_w\) is the contraction amplitude, \(\sigma_w\) 
controls the spatial extent, and \(f_w(t)\) is a temporal fade-in function to reduce discontinuity as the wave reaches the end of the mesh. The deformed vertex position at time \(t\) is then defined as:
\begin{equation}
\mathbf{x}_i(t) = \mathbf{c}(s_i) + \left(1 - \beta_i(t)\right) \mathbf{r}_i.
\end{equation}

where $\beta$ is the total contraction at vertex i and time t, defined as the sum of all traveling gaussian waves $\beta_i(t) = \sum_w g_{i,w}(t)$.
\paragraph{Centerline shift.} We simulate additional anatomical deformation by transforming the centerline curve and propagating these changes to the mesh surface. We apply smooth transformations to the centerline, including rigid shifts, linear translations with fixed endpoints, sinusoidal warps, or exponential tail bends while keeping the radial offset of each vertex constant. Additionally, the centerline transformation framework is extensible, any smooth deformation function can be applied to simulate arbitrary anatomical variation.

\subsection{Sim2real translation}
\paragraph{Problem setup.}
Given a synthetic colonoscopy depth sequence $d_{1:T}^{s}$ rendered from a deforming C3VD/C3VDv2 mesh, a text prompt $c$, and a single real colonoscopy appearance anchor $x_{\mathrm{ref}}^{r}$, our goal is to generate a video $\hat{x}_{1:T}$ that preserves the geometry of $d_{1:T}^{s}$ while matching the appearance distribution of in vivo colonoscopy. We address this with a two-stage adaptation of LTX-2.3 \cite{hacohen2026ltx}, a latent video diffusion-transformer trained with a rectified-flow objective \cite{lipman2022flow}. All pretrained backbone weights are kept frozen, and only low-rank adapters are trained.

\paragraph{Stage 1: real-domain video adaptation.}
First, we adapt the pretrained LTX-2.3 video prior to the real colonoscopy domain using RGB-only LoRA fine-tuning on EndoMapper videos. This stage exposes the model to in vivo mucosal texture, illumination, specularities, endoscopic camera motion, and temporal appearance variation without imposing synthetic geometric conditioning. Each clip is paired with a short text caption, such as ``real colonoscopy video, white-light imaging.'' After training, the Stage 1 LoRA is merged into the frozen backbone, producing a real-domain video generator $\theta^{(1)}$. This design is important in our low-data setting: our training corpus contains approximately 2k informative EndoMapper clips, which is small relative to the broad video distribution used to pretrain LTX-2.3.

\paragraph{Stage 2: depth-conditioned IC-LoRA.}
Second, we train a depth-conditioned in-context LoRA \cite{huang2024context} on top of $\theta^{(1)}$. Because real colonoscopy videos do not provide ground-truth depth, we estimate pseudo-depth for EndoMapper clips using ColonCrafter \cite{hardy2025coloncrafter}. During training, the real RGB clip is the target video, and the corresponding pseudo-depth clip is used as the conditioning stream. This teaches the model to use dense depth-like structure while keeping the generated target distribution in the real in vivo domain. At inference, the pseudo-depth input is replaced by rendered synthetic depth from deforming C3VD/C3VDv2 meshes.

We encode the depth conditioning clip with the same latent video autoencoder as the RGB video and provide it to the denoising transformer as an IC-LoRA reference latent. The loss is applied only to the generated target video latents. To reduce the domain gap between monocular pseudo-depth and rendered metric depth, we apply lightweight augmentations to the depth stream during training, including gamma jitter and percentile stretching.

\paragraph{Training objective.}
Both stages use the LTX-2.3 rectified-flow velocity-prediction objective \cite{lipman2022flow}. Given a clean target latent $z_0$, noise $z_1$, timestep $t$, text condition $c$, and interpolated latent $z_t$, the model predicts the velocity target $u_t$. Stage 2 adds encoded depth conditioning $E(\tilde{d})$:
\begin{equation}
\mathcal{L}_{\mathrm{IC}}
=
\mathbb{E}_{z_0,z_1,t,c,\tilde{d}}
\left[
\left\|
v_\theta(z_t,E(\tilde{d}),t,\tau(c)) - u_t
\right\|_2^2
\right],
\end{equation}
where $\tilde{d}$ denotes the augmented pseudo-depth clip, $E(\cdot)$ is the video autoencoder, and $\tau(\cdot)$ is the text encoder.

\paragraph{Sim-to-real inference.}
At inference, the model receives a rendered synthetic depth sequence, a real EndoMapper reference frame, and the prompt ``real colonoscopy image.'' The reference frame anchors color, texture, and illumination, while the synthetic depth sequence provides the primary geometric control. To reduce the train/test gap between ColonCrafter pseudo-depth and rendered metric depth, we apply lightweight inference-time depth preprocessing before conditioning.

\begin{table}[t]
\centering
\caption{
Dataset validation: visual realism and geometric consistency.
Realism metrics compare generated videos to real in-vivo colonoscopy videos.
Geometric metrics evaluate alignment between generated RGB and the conditioning geometry.
}
\label{tab:realism_geometry}
\small
\setlength{\tabcolsep}{5pt}
\renewcommand{\arraystretch}{1.05}
\begin{tabular}{lcccccc}
\toprule
& \multicolumn{3}{c}{\textbf{Visual realism}} 
& \multicolumn{3}{c}{\textbf{Geometric consistency}} \\
\cmidrule(lr){2-4}\cmidrule(lr){5-7}
\textbf{Method} 
& FID $\downarrow$ 
& KID $\downarrow$ 
& FVD $\downarrow$ 
& HF Dice $\uparrow$ 
& HF Chamfer $\downarrow$ 
& Depth $\rho$ $\uparrow$ \\
\midrule
Pix2Pix 
& 214.31 & 0.21 & 3597.76 
& 0.96 & 0.09 & \textbf{0.66} \\

AnimateDiff 
& 143.45 & 0.15 & 3138.80
& 0.95 & 0.63 & 0.13 \\

\textbf{C3VD-DefCol} 
& \textbf{96.32} & \textbf{0.07} & \textbf{2434.79} 
& \textbf{0.97} & \textbf{0.08} & 0.64 \\
\bottomrule
\end{tabular}
\end{table}
\section{Experimental Details}

\textbf{Training data and configuration.} We train on EndoMapper in vivo colonoscopy videos after filtering blurred, `dirty' lens, and uninformative frames. Videos are sampled into 49-frame clips at 16 fps and resized to $512 \times 512$, yielding 2066 training clips and 114 held-out clips. ColonCrafter pseudo-depth maps are computed for the same clips and used as the conditioning stream for Stage 2. Both stages adapt the open-source LTX-2.3-22B checkpoint using rank/alpha $=32/32$ LoRA adapters, AdamW, mixed-precision training, and frozen backbone weights. Stage 1 trains an RGB-only LoRA on EndoMapper videos; Stage 2 trains the depth IC-LoRA using EndoMapper RGB clips paired with pseudo-depth. All LoRA training and sim-to-real generation were performed on NVIDIA H100 GPUs with 80 GB memory. Stage 1 was trained for 8250 steps, Stage 2 was trained for 3000 steps.

\textbf{Inference and evaluation.} For evaluation, we generate sim-to-real videos from rendered C3VD/C3VDv2 deforming depth sequences at $512 \times 512$ resolution and 16 fps. Each sequence is generated for 89 frames using video-to-video conditioning with a manually selected EndoMapper frame as the appearance anchor. Since the first few frames can retain residual geometry from the reference image before the depth control fully dominates, we discard the first 5 frames and evaluate the remaining frames for all generated videos. Before conditioning, rendered depth maps are normalized and lightly blurred to better match ColonCrafter pseudo-depth statistics. For `down-the-barrel' views with large far-depth regions, we additionally use percentile-stretched inverse-depth normalization followed by gamma remapping to reduce conditioning leakage into RGB appearance. Generated videos are evaluated for visual realism, geometric consistency with the input depth, temporal consistency, and downstream 3D reconstruction performance.

\section{Experiments}
\subsection{Dataset validation}
We validate C3VD-DEFCOL along three axes required for deformable colonoscopy reconstruction benchmarks: visual realism, geometric consistency, and temporal consistency. We compare our geometry-conditioned LTX video generation model against Pix2Pix \cite{isola2017image}, a frame-wise image-to-image translation baseline, and AnimateDiff \cite{guo2023animatediff}, a video diffusion baseline using Stable Diffusion \cite{rombach2021high} with ControlNet\cite{zhang2023adding}.

\paragraph{Visual realism.}
We evaluate visual realism using FID \cite{heusel2017gans}, KID \cite{binkowski2018demystifying}, and FVD \cite{unterthiner2018towards} by comparing generated videos against real in-vivo colonoscopy videos as shown in \autoref{tab:realism_geometry}. FID and KID measure frame-level distributional similarity, while FVD measures video-level similarity in a spatiotemporal feature space. These metrics do not certify clinical realism, but they provide a standardized comparison of whether sim-to-real translation moves rendered colonoscopy videos closer to the real video distribution. As shown in Table~\ref{tab:realism_geometry}, C3VD-DEFCOL substantially improves over the frame-wise Pix2Pix and AnimateDiff baselines across all realism metrics, suggesting that LTX video diffusion better captures the color, texture, illumination, and temporal appearance statistics of in-vivo colonoscopy.

\begin{figure}
    \centering
    \includegraphics[width=0.55\linewidth]{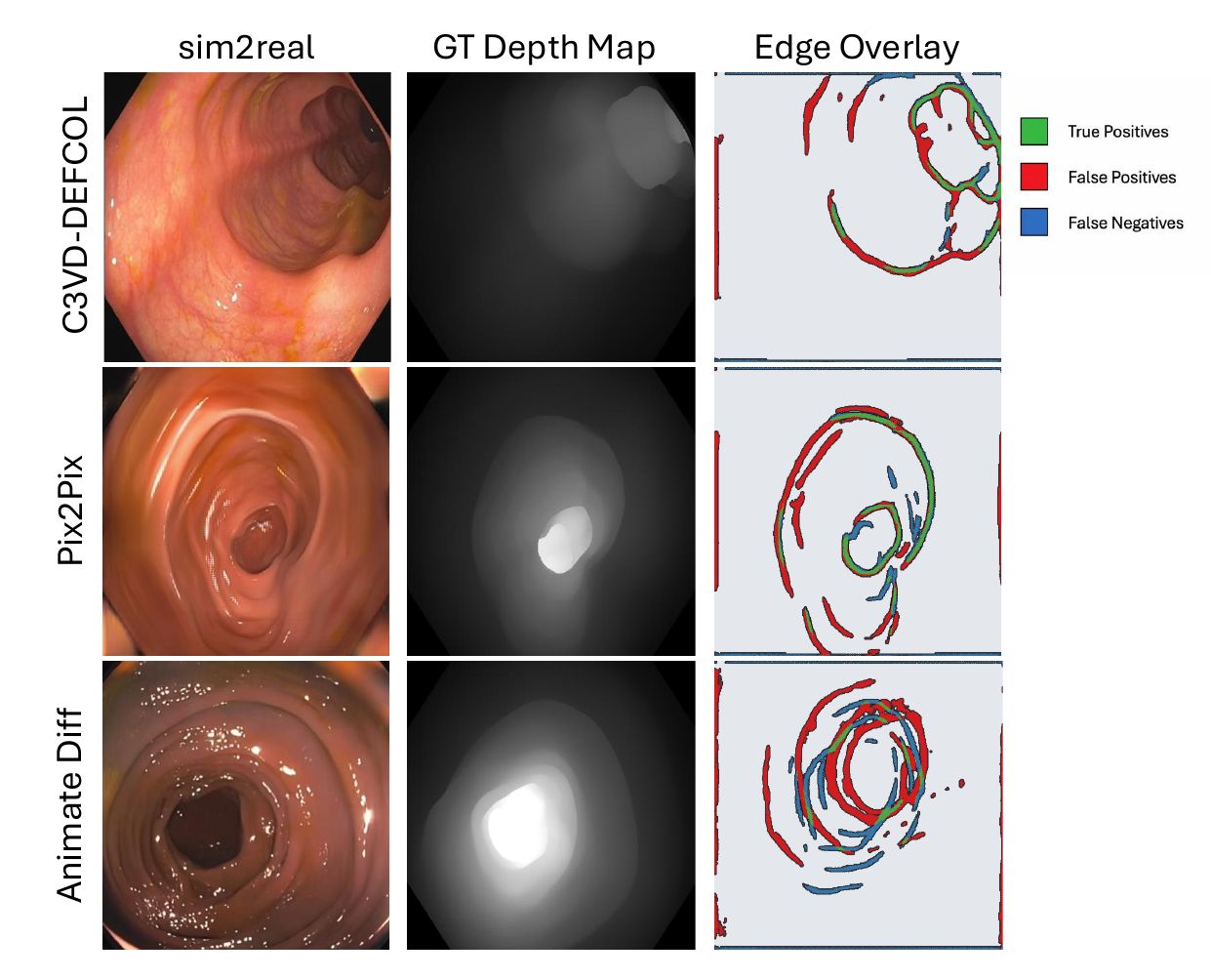}
    \caption{\textbf{Qualitative comparison of geometric consistency.} Edges extracted from depth maps and generated RGB images are overlaid to assess haustral-fold alignment. The strong overlap shows that C3VD-DEFCOL generates RGB images that are highly consistent with the input depth map.}
    \label{fig:geo_consistency}
\end{figure}

\paragraph{Geometric consistency.}
We verify whether the generated RGB remains aligned with the conditioning geometry using haustral-fold alignment and depth-order consistency as shown in Figure \ref{fig:geo_consistency}. For haustral-fold alignment, we compare fold (edge) maps derived from the input depth using DexiNed \cite{xsoria2020dexined} with fold maps extracted from the generated RGB using HalF-SAM \cite{golhar2025half}, then report Dice overlap and Chamfer distance. This gives a colonoscopy-specific measure of whether important anatomical landmarks remain spatially aligned after sim-to-real translation. We also estimate pseudo-depth from generated RGB using ColonCrafter and compare it with the rendered ground-truth depth using Spearman rank correlation, since ColonCrafter predicts relative depth while C3VD-DEFCOL provides metric depth. C3VD-DEFCOL achieves strong fold alignment and comparable depth-order consistency to Pix2Pix, indicating that the generated appearance remains geometrically faithful enough for quantitative reconstruction evaluation.

\begin{table}[t]
\centering
\caption{
Temporal consistency evaluation using CD-FVD and feature-track persistence.
Higher track persistence and larger feature counts indicate more stable and matchable appearance over time.
}
\label{tab:temporal_consistency}
\small
\setlength{\tabcolsep}{4pt}
\renewcommand{\arraystretch}{1.05}
\begin{tabular}{lccccccc}
\toprule
& \multirow{2}{*}{\textbf{CD-FVD $\downarrow$}} 
& \multicolumn{3}{c}{\textbf{SG tracks}} 
& \multicolumn{3}{c}{\textbf{SIFT tracks}} \\
\cmidrule(lr){3-5}\cmidrule(lr){6-8}
\textbf{Method} 
& 
& \# feat. $\uparrow$ 
& $\geq 5$ $\uparrow$ 
& $\geq 10$ $\uparrow$ 
& \# feat. $\uparrow$ 
& $\geq 5$ $\uparrow$ 
& $\geq 10$ $\uparrow$ \\
\midrule
Pix2Pix     
& 1033.82 
& 72.8 & 0.220 & 0.060 
& 17.1 & 0.171 & 0.061 \\

AnimateDiff 
& 1004.24 
& \textbf{152.9} & 0.079 & 0.007 
& 0.5 & 0.007 & 0.000 \\

\textbf{C3VD-DefCol} 
& \textbf{373.17} 
& 92.0 & \textbf{0.366} & \textbf{0.151} 
& \textbf{20.9} & \textbf{0.259} & \textbf{0.092} \\
\bottomrule
\end{tabular}
\end{table}
\paragraph{Temporal consistency.}
We evaluate temporal consistency using CD-FVD \cite{ge2024content} and feature-track persistence. CD-FVD is a content-debiased variant of FVD that reduces sensitivity to static frame appearance and better reflects temporal coherence. Feature tracking directly tests whether generated structures remain matchable over time: we track features using SIFT \cite{lowe2004distinctive} and SuperPoint/SuperGlue (SG)-based matching \cite{sarlin2020superglue}, then report the average number of detected features and the fraction of tracks lasting at least 5 and 10 frames in \autoref{tab:temporal_consistency}. Higher values indicate more stable appearance, which is important because SLAM and SfM rely on persistent correspondences. C3VD-DEFCOL improves CD-FVD and produces longer-lived tracks than Pix2Pix. The lower absolute SIFT persistence is expected because colonoscopy images are often texture-poor, specular, and repetitive. Interestingly, AnimateDiff produced the highest average feature count, but many of these features originated from specular reflection points. As a result, despite the large number of detected features, they were often short-lived, consistent with the transient nature of specular highlights. 
\begin{table}
    \centering
    \caption{COLMAP performance across different datasets, regions, and deformation levels.}
    \label{tab:colmap_performance}
    \small
    \setlength{\tabcolsep}{5pt}
    \begin{tabular}{llcccc}
        \toprule
         & & \textbf{Rigid} & \textbf{Low def.} & \textbf{Medium def.} & \textbf{High def.} \\
        \cmidrule(lr){3-3} \cmidrule(lr){4-4} \cmidrule(lr){5-5} \cmidrule(lr){6-6}
        \textbf{Dataset} & \textbf{Region} & \textbf{RMSE ATE} & \textbf{RMSE ATE} & \textbf{RMSE ATE} & \textbf{RMSE ATE} \\
         & & [mm]$\downarrow$ & [mm]$\downarrow$ & [mm]$\downarrow$ & [mm]$\downarrow$ \\
        \midrule
        C3VD & Cecum & 0.32 & -- & -- & -- \\
        \midrule
        \multirow{3}{*}{C3VD-DEFCOL} & Ascending  & 0.40 & 1.59 & 3.34 & 5.23 \\
                                     & Transverse & 0.28 & 0.52 & 1.08 & 1.23 \\
                                     & Descending & 0.68 & 2.17 & 2.13 & 7.00 \\
        \bottomrule
    \end{tabular}
\end{table}

\begin{figure}
    \centering
    \includegraphics[width=0.7\linewidth]{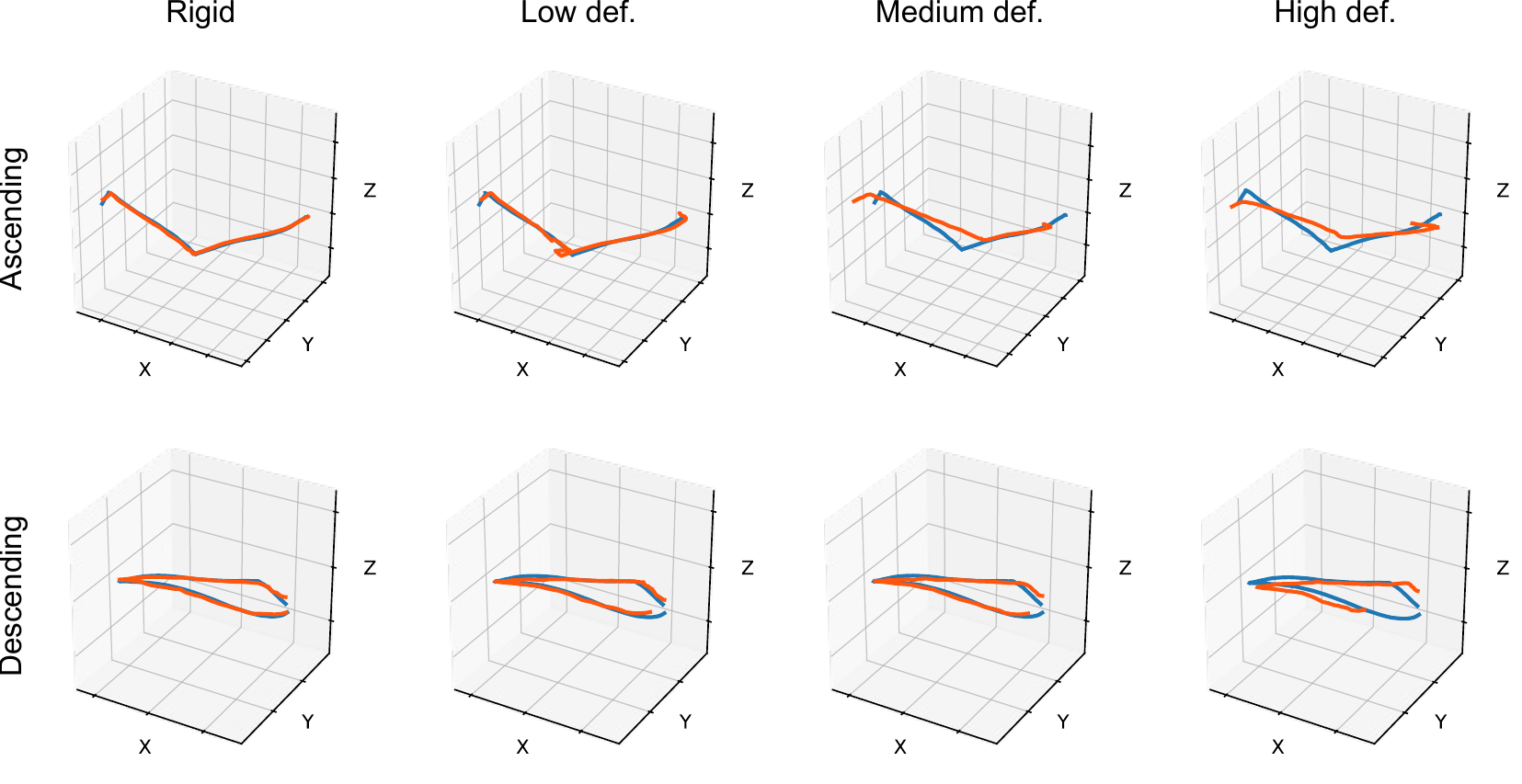}
    \caption{Estimated 3D trajectories from COLMAP (orange) compared with the ground truth (blue) for different deformation levels across two sequences from the C3VD-DEFCOL dataset.}
    \label{fig:colmap_traj}
\end{figure}

\subsection{Pose Estimation}

Pose estimation in intraoperative colonoscopy is an actively researched topic that could enable assisted and robotic procedures. We selected COLMAP \cite{schonberger2016structure} as a widely recognized and trusted baseline to assess the value of the proposed benchmark for the specific task of 3D reconstruction.

\begin{table}
    \centering
    \caption{COLMAP performance evaluation using sim-to-real textures from different reference frames.}
    \label{tab:colmap_texture_comparison}
    \small
    \setlength{\tabcolsep}{12pt} 
    \begin{tabular}{lccc}
        \toprule
         & \textbf{Low def.} & \textbf{Medium def.} & \textbf{High def.} \\
        \cmidrule(lr){2-2} \cmidrule(lr){3-3} \cmidrule(lr){4-4}
        \textbf{Texture Source} & \textbf{RMSE ATE} & \textbf{RMSE ATE} & \textbf{RMSE ATE} \\
        \textbf{(Ref. Frame)} & [mm]$\downarrow$ & [mm]$\downarrow$ & [mm]$\downarrow$ \\
        \midrule
        Reference Frame A & 1.75 & 4.62 & 4.47 \\
        Reference Frame B & 1.49 & 4.14 & 5.72 \\
        \bottomrule
    \end{tabular}
\end{table}

\autoref{tab:colmap_performance} shows COLMAP results for rigid sequences across the original C3VD and the proposed C3VD-DEFCOL datasets. The trajectory error (ATE), after \textit{Sim(3)} alignment, is comparable in both datasets. This further confirms that the sim-to-real texture mapping is consistent with the underlying 3D geometry, ensuring that the pseudo-real textures do not introduce artifacts that would otherwise mislead feature matching.
Evaluation of non-rigid sequences in C3VD-DEFCOL reveals a clear correlation between deformation intensity and localization error. As deformation progresses from Low to High, the ATE increases monotonically. 

This behavior can also be qualitatively observed in \autoref{fig:colmap_traj}, where the COLMAP trajectories progressively diverge from the ground truth as the deformation level increases. This trend is consistent with the theoretical constraints of rigid-body SfM methods, where non-rigid surface fluctuations are interpreted as noise or outliers, leading to reduced trajectory accuracy.

Feature-based methods such as COLMAP depend on texture and often struggle to reconstruct poorly textured regions. In \autoref{tab:colmap_texture_comparison}, we compare the trajectory error for two versions of the same sequence, both depicting the exact same camera motion and surface geometry but with different texture reference frames. The results are consistent, suggesting that, given fixed geometry, the sim-to-real style transfer provides a degree of texture invariance. 

\paragraph{Limitations}
The geometric deformation models, while controllable, do not faithfully replicate the biomechanical properties of colonic tissue. In particular, the dataset does not capture non-rigid effects arising from insufflation, tissue-tool interaction, tissue elasticity, and related physiological factors. The sim2real videos may also contain generative artifacts, including hallucinated structures, color shifts, or occasional appearance-geometry mismatch. The source geometries and trajectories do not capture the full diversity of real colonoscopy and also do not include bleeding, inflammation, or poor bowel preparation artifacts. Hence, performance improvements on C3VD-DEFCOL should be interpreted as evidence of robustness in the controlled deformation setting and may not generalize to robust clinical performance. Future work should expand the range of biomechanical deformation models, appearance conditions, anatomical variation, and validation against larger-scale in vivo datasets.

\section{Conclusion}
We presented C3VD-DEFCOL, a deformable colonoscopy video dataset and evaluation framework that pairs realistic sim-to-real appearance with dense, time-resolved geometric ground truth. The dataset provides pixel-wise 3D ground truth, camera poses, and deforming mesh annotations across controlled deformation modes and magnitudes. By combining parameterized tissue motion, dense labels, and realistic video appearance, C3VD-DEFCOL addresses a key gap in existing colonoscopy benchmarks: the lack of data that is both visually realistic and fully measurable for evaluating algorithms under non-rigid deformation.
We evaluate the dataset along three axes, visual realism, geometric consistency, and temporal consistency, and demonstrate its benchmarking utility through a downstream SLAM-based pose estimation study, where performance degradation can be quantified as deformation magnitude increases. Beyond the released dataset, C3VD-DEFCOL is an extensible framework: researchers can generate tailored videos by varying mesh geometry, deformation type and magnitude, camera trajectory, and appearance, while newer sim-to-real models can be swapped into the pipeline as they improve. We hope this resource enables more rigorous training and testing of colonoscopy 3D reconstruction methods and helps clarify the conditions under which these methods succeed or fail.

\medskip

{
\small
\bibliographystyle{unsrt}
\bibliography{egbib}


}



\newpage

\end{document}